\documentclass[sigconf]{acmart}

\usepackage{amsmath}
\usepackage{multirow}
\usepackage{microtype}
\usepackage{graphicx}
\usepackage{booktabs}
\usepackage{xspace}
\usepackage{subcaption}
\usepackage{balance}

\DeclareMathOperator*{\argmax}{argmax}

\AtBeginDocument{%
  \providecommand\BibTeX{{%
    \normalfont B\kern-0.5em{\scshape i\kern-0.25em b}\kern-0.8em\TeX}}}

\setcopyright{acmcopyright}
\copyrightyear{2021}
\acmYear{2021}
\acmDOI{10.1145/1122445.1122456}

\acmConference[WSDM'22]{WSDM'22: The 15th ACM International Conference on Web Search and Data Mining}{Feb. 21--25, 2022}{Phoenix, Arizona}
\acmBooktitle{WSDM'22: The 15th ACM International Conference on Web Search and Data Mining, Feb. 21--25, 2022, Phoenix, Arizona}
\acmPrice{15.00}
\acmISBN{978-1-4503-XXXX-X/18/06}

\newcommand{\modelname}{IMEA\xspace}

\copyrightyear{2022}
\acmYear{2022}
\setcopyright{acmcopyright}\acmConference[WSDM '22]{Proceedings of the Fifteenth ACM International Conference on Web Search and Data Mining}{February 21--25, 2022}{Tempe, AZ, USA}
\acmBooktitle{Proceedings of the Fifteenth ACM International Conference on Web Search and Data Mining (WSDM '22), February 21--25, 2022, Tempe, AZ, USA}
\acmPrice{15.00}
\acmDOI{10.1145/3488560.3498523}
\acmISBN{978-1-4503-9132-0/22/02}

\begin{document}
\fancyhead{}

\title{Informed Multi-context Entity Alignment}

\author{Kexuan Xin$^{1}$, Zequn Sun$^2$, Wen Hua$^{1*}$, Wei Hu$^{2*}$, Xiaofang Zhou$^{3}$}\thanks{$^*$Corresponding authors}
\affiliation{
\institution{$^1$The University of Queensland, Brisbane, QLD 4072, Australia} 
\institution{$^2$State Key Laboratory for Novel Software Technology, Nanjing University, China}
\institution{$^3$Hong Kong University of Science and Technology, Kowloon, Hong Kong}}
\email{{uqkxin, w.hua}@uq.edu.au, zqsun.nju@gmail.com, whu@nju.edu.cn, zxf@cse.ust.hk}

\renewcommand{\shortauthors}{Kexuan Xin and Zequn Sun, et al.}

\begin{abstract}
Entity alignment is a crucial step in integrating knowledge graphs (KGs) from multiple sources. 
Previous attempts at entity alignment have explored different KG structures, such as neighborhood-based and path-based contexts, to learn entity embeddings, but they are limited in capturing the multi-context features. 
Moreover, most approaches directly utilize the embedding similarity to determine entity alignment without considering the global interaction among entities and relations. 
In this work, we propose an Informed Multi-context Entity Alignment (IMEA) model to address these issues.
In particular, we introduce Transformer to flexibly capture the relation, path, and neighborhood contexts, and design holistic reasoning to estimate alignment probabilities based on both embedding similarity and the relation/entity functionality. The alignment evidence obtained from holistic reasoning is further injected back into the Transformer via the proposed soft label editing to inform embedding learning.
Experimental results on several benchmark datasets demonstrate the superiority of our IMEA model compared with existing state-of-the-art entity alignment methods.
\end{abstract}



\maketitle

\section{Introduction}
Knowledge graphs (KGs) possess a large number of facts about the real world in a structured manner. 
Each fact consists of two entities and their relation in the form of a triple like \textit{(subject entity, relation, object entity)}. 
KGs are becoming increasingly important because they can simplify knowledge acquisition, storage, and reasoning, and successfully support many knowledge-powered tasks such as semantic search and recommendation. 
As different KGs are constructed by different people for different purposes using different data sources,
they are far from complete, yet complementary to each other. 
Knowledge from a single KG usually cannot satisfy the requirement of downstream tasks, calling for the integration of multiple KGs for knowledge fusion.
Entity alignment, which aims to identify equivalent entities from different KGs, is a fundamental task in KG fusion.
In recent years, there has been an increasing interest in embedding-based entity alignment \cite{chen2017multilingual,wang2018cross,cao2019multi,sun2020knowledge}. 
It encodes entities into vector space for similarity calculation, such that entity alignment can be achieved via nearest neighbor search.
However, current embedding-based models still face two major challenges regarding alignment learning and inference.

\textbf{Challenge 1:} How to effectively represent entities in the alignment learning phase?
Existing embedding-based methods do not fully explore the inherent structures of KGs for entity embedding. 
Most of them only utilize a one-sided view of the structures, such as relation triples \cite{chen2017multilingual,sun2017cross}, paths \cite{zhu2017iterative,guo2019learning}, 
and neighborhood subgraphs \cite{wang2018cross, sun2020knowledge, cao2019multi, wu2019relation, wu2019jointly, li2019semi, wu2020neighborhood, xu2019cross, yang2019aligning}.
Intuitively, all these multi-context views could be helpful for embedding learning and provide additional evidence for entity alignment.
Multi-view learning \cite{zhang2019multi} is not the best solution to aggregate these features as it requires manual feature selection to determine useful views. 
Besides, KGs are heterogeneous, indicating that not all features in a helpful view could contribute to alignment learning, such as the noisy neighbors in neighborhood subgraphs \cite{sun2020knowledge}. 
Hence, it is necessary to flexibly and selectively incorporate different types of structural contexts for entity embedding.

\textbf{Challenge 2:} How to make full use of the learned embeddings and achieve informed alignment learning? Existing models directly rely on entity embeddings to estimate their semantic similarity, which is used for alignment decisions while ignoring the correlation between entities and relations in the KG. Although some studies \cite{wu2019jointly, zhu2021relation} have tried to conduct joint training of entity and relation to improve the quality of the learned embeddings, alignment of each entity is still conducted independently from a local perspective. In fact, entity alignment is a holistic process: the neighboring entities and relations of the aligned pair should also be equivalent. The methods \cite{zeng2020collective,zeng2021reinforcement} conduct entity alignment from a global perspective by modelling it as a stable matching problem. Whereas, such reasoning is regarded as a post-processing step without utilizing the holistic information to inform the embedding learning module. We argue that holistic inference can provide crucial clues to achieve reliable entity alignment, and the inferred knowledge can offer reliable alignment evidence and help to build and strengthen the trust towards black-box embedding-based entity alignment.

\textbf{Contributions:} In this work, we introduce \textbf{\modelname}, an \textbf{I}nformed \textbf{M}ulti-context \textbf{E}ntity \textbf{A}lignment model, to tackle the above obstacles.
Motivated by the strong expressiveness and representation learning ability of Transformer \cite{vaswani2017attention}, 
we design a novel Transformer architecture to encode the multiple structural contexts in a KG while capturing the deep alignment interaction across different KGs.
Specifically, we apply two Transformers to encode the neighborhood subgraphs and entity paths, respectively.
To capture the semantics of relation triples, we introduce relation regularization based on TransE \cite{bordes2013translating} at the input layer.
We generate entity paths by random walks across two KGs. 
To capture the deep alignment interactions, we replace the entity in a path with its counterpart in the seed alignment (i.e., training data) to generate a new semantically equivalent path.
In this way, a path may contain entities from the two KGs that remain to be aligned, 
and the Transformer is trained to fit such a cross-KG entity sequence and its equivalent path.
The two Transformers share parameters.
Their training objectives are central entity prediction and mask entity prediction, respectively.
The self-attention mechanism of Transformer can recognize and highlight helpful features in an end-to-end manner. 

Furthermore, we design a holistic reasoning process for entity alignment, to complement the imperfection of embedding similarities.
It compares entities by reasoning over their neighborhood subgraphs based on both embedding similarities and the holistic relation and entity functionality (i.e., the importance for identifying similar entities and relations).
The inferred alignment probabilities are injected back into the Transformer as soft labels to assist alignment learning. In this way, both the positive and negative alignment evidence obtained from the holistic inference can be utilized to adjust soft labels and guide embedding learning, thus improving the overall performance and reliability of entity alignment.

We conduct extensive experiments including ablation studies and case studies on the benchmark dataset OpenEA \cite{sun13benchmarking} to evaluate the effectiveness of our \modelname model. Experimental results demonstrate that \modelname can achieve the state-of-the-art performance by combining multi-context Transformer and holistic reasoning,
and the global evidence is quite effective in informing alignment learning.

\section{Related Work}
\textbf{KG Contexts for Entity Alignment.} 
Existing structure-based entity alignment methods utilize different types of structural contexts to generate representative KG embeddings. 
A line of studies learns from \textit{relation triples} and encodes local semantic information of both entities and relations. 
TransE \cite{bordes2013translating} is one of the most fundamental embedding techniques that has been adopted in a series of entity alignment models such as MTransE \cite{chen2017multilingual}, JAPE \cite{sun2017cross}, BootEA \cite{sun2018bootstrapping} and SEA \cite{pei2019semi}.
Inspired by the success of graph neural networks (GNNs) like GCN \cite{kipf2016semi} and GAT \cite{velickovic2018graph} in graph embedding learning,
recent entity alignment studies incorporate various GNN variants to encode \textit{subgraph contexts} for neighborhood-aware KG embeddings. 
For example, GCN-Align \cite{wang2018cross} employs vanilla GCN for KG structure modeling, 
but suffers from the heterogeneity of different KGs. 
To address this issue, MuGNN \cite{cao2019multi} preprocesses KGs by KG completion and applies an attention mechanism to highlight salient neighbors.
Other work improves neighborhood-aware entity alignment by introducing relation-aware GNNs \cite{li2019semi,ye2019vectorized,yu2020generalized,mao2020mraea,mao2020relational}, multi-hop GNNs \cite{sun2020knowledge,mao2021boosting} and hyperbolic GNNs \cite{sun2020knowledge}.
Besides the triple and neighbourhood context, 
RSN4EA \cite{guo2019learning} and IPTransE \cite{zhu2017iterative} further explore long-term dependency along \textit{relation paths} in KGs. 
There are also some entity alignment models that consider additional information such as \textit{entity names} \cite{xu2019cross,wu2019relation,wu2019jointly,wu2020neighborhood}, \textit{entity attributes} \cite{trisedya2019entity,zhang2019multi,liu2020exploring} and \textit{textual descriptions} \cite{chen2018co}.
Although structural contexts are abundant and always available in KGs, few existing EA models have tried to integrate them into the embedding learning process. In this work, we introduce Transformer to capture multi-context interactions and learn more comprehensive entity embeddings.

\noindent\textbf{Collective Entity Alignment.}
Entities are usually aligned independently in existing EA models. Only two methods have attempted to achieve collective entity alignment.
CEAFF \cite{zeng2020collective, zeng2021reinforcement} formulates alignment inference as a stable matching problem. 
RMN \cite{zhu2021relation} designs an iterative framework to leverage positive interactions between entity and relation alignment. However, it can only capture the interactions from seed entity alignment and requires extra pre-aligned relations, which is not always feasible in practice. Furthermore, both methods treat alignment learning and alignment inference separately, e.g., by regarding holistic reasoning as a post-processing step to adjust the alignment results \cite{zeng2020collective}. By contrast, our \modelname model can inject the holistic knowledge obtained during alignment inference into the Transformer via soft label editing, so as to guide the embedding learning process.

\section{Methodology}
\subsection{Framework}
We define a KG as $\mathcal{G}=(\mathcal{E},\mathcal{R},\mathcal{T})$, 
where $\mathcal{E}$ and $\mathcal{R}$ denote the sets of entities and relations, respectively, and $\mathcal{T}\subset\mathcal{E}\times\mathcal{R}\times\mathcal{E}$ is the set of relation triples. Each triple can be represented as $(s, r, o)$, where $s\in\mathcal{E}$ and $o\in\mathcal{E}$ denote the subject (head entity) and object (tail entity), respectively, and $r\in\mathcal{R}$ is the relation.
Given two heterogeneous KGs, i.e., a source KG $\mathcal{G}$ and a target KG $\mathcal{G}'$, entity alignment (EA) aims to find as many aligned entity pairs as possible $\mathcal{A} = \{(e, e') \in \mathcal{E} \times \mathcal{E}'| e \equiv e'\}$, 
where the equivalent relationship $\equiv$ indicates that entities $e$ and $e'$ from different KGs represent the same real-world identity. 
Typically, to jump-start alignment learning, some seed entity alignment $\mathcal{A}^s$ is provided as training data.

\begin{figure*}[!t]
    \centering
    \includegraphics[width=0.68\linewidth]{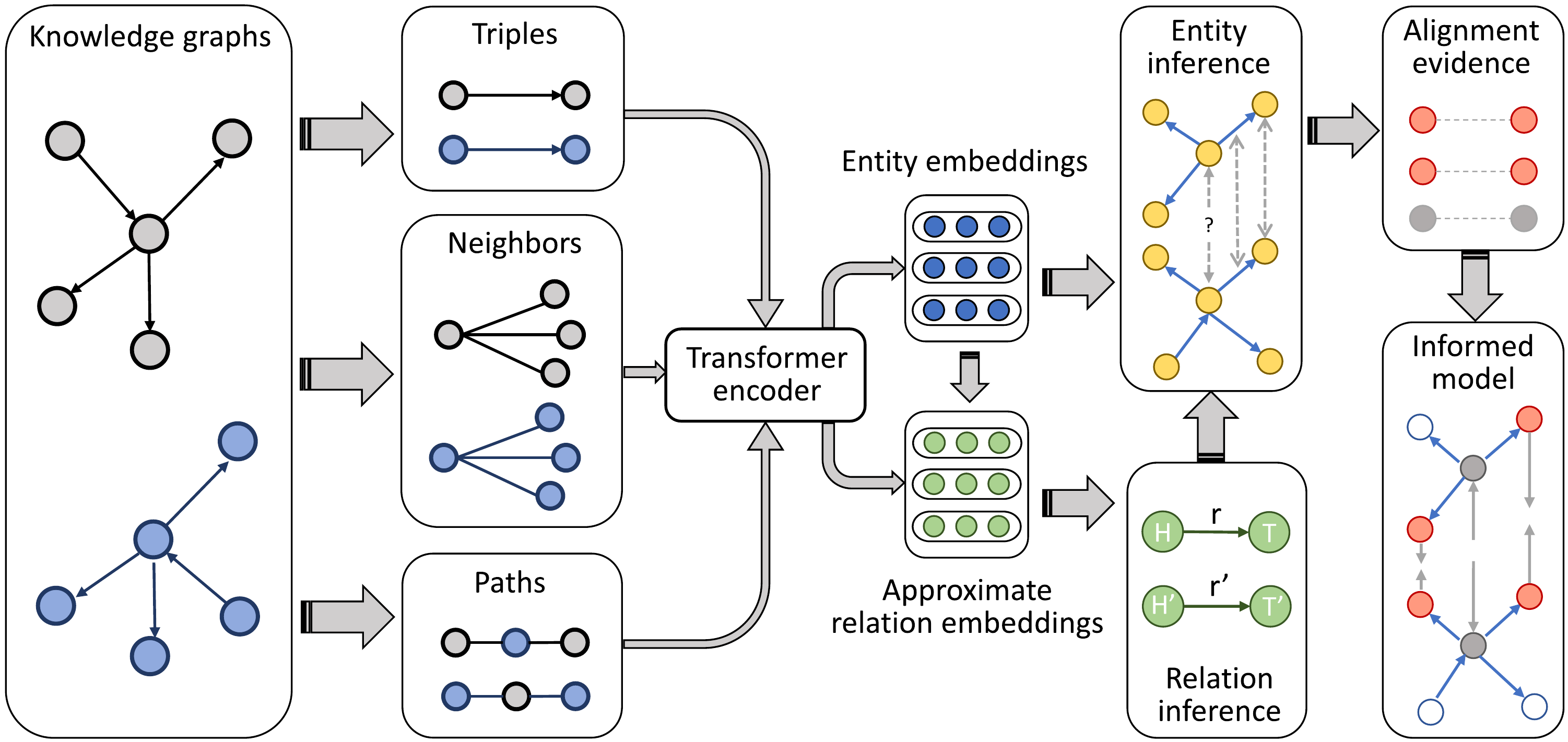}
    \caption{\label{fig:framework.pdf}Framework of the proposed informed multi-context entity alignment model.}
\end{figure*}

Figure \ref{fig:framework.pdf} shows the overall framework of our \modelname (Informed Multi-context Entity Alignment) model, which contains three main components:
Firstly, we introduce Transformer as the contextualized encoder to capture the multiple structural contexts (i.e., triples, neighborhood subgraphs and paths) and generate representative KG embeddings for alignment learning. 
Next, we identify candidate target entities for each source entity based on their embedding similarity and perform holistic reasoning for each candidate entity pair. 
Finally, we inform the embedding learning module by injecting the alignment evidence into the training loss using our proposed soft label editing strategy. 
We will introduce each component in detail in the following sections.

\subsection{Multi-context Transformer}
\begin{figure*}
    \centering
    \includegraphics[width=0.8\linewidth]{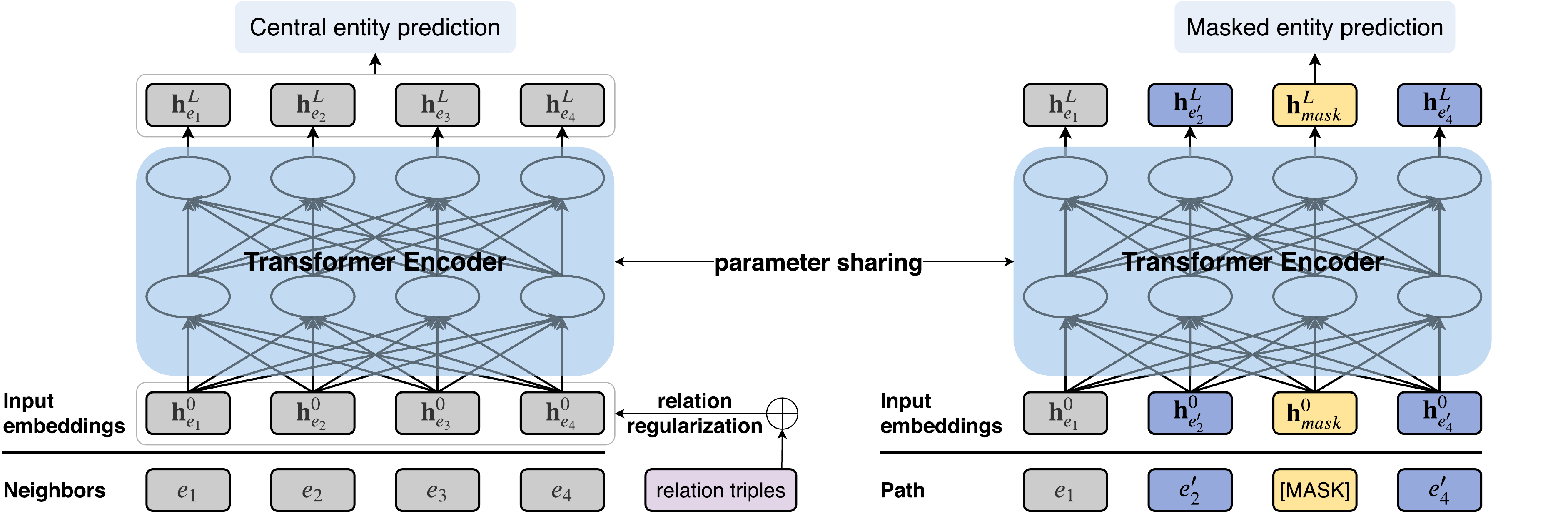}
    \caption{\label{fig:transformer}Illustration of multi-context Transformer.}
\end{figure*}

Figure \ref{fig:transformer} presents the architecture of our multi-context transformer. 
Specifically, we apply two Transformers to encode the neighborhood context and path context, respectively.
The neighborhood context is generated by sampling neighboring entities. 
To capture the relation semantics, we introduce a relation regularization based on the translational embedding function \cite{bordes2013translating} at the input layer.
The path context is represented as an entity sequence generated by random walks. 
To capture the deep alignment interactions across different KGs, 
we replace the entity in a path with its counterpart in the seed entity alignment to generate a new semantically equivalent path.
In this way, a path may contain entities from the two KGs that remain to be aligned, 
and the Transformer seeks to fit such a cross-KG entity sequence and its equivalent path to capture the alignment information between two KGs.

\subsubsection{Intra-KG Neighbourhood Encoding}
Neighborhood information is critical in representing KG structures for entity alignment, as similar entities usually have similar neighborhood subgraphs. 
Considering that an entity may have a very large number of neighbors, 
we repeatedly sample $n$ neighbors of each entity during training and feed these neighbors into Transformer to capture their interactions. 
Specifically, given an entity $e$ and its sampled neighbours $(e_1, e_2,..., e_n)$, 
we regard the neighbourhood as a sequence and feed it into a Transformer with $L$ self-attention layers stacked. 
The input embeddings of Transformer are $(\mathbf{h}_{e_1}^0, \mathbf{h}_{e_2}^0, \cdots, \mathbf{h}_{e_n}^0)$, 
and the output of the $L$-th layer is generated as follows:
\begin{equation}
    \mathbf{h}_{e_1}^L, \mathbf{h}_{e_2}^L, \cdots, \mathbf{h}_{e_n}^L = \text{Encoder}(\mathbf{h}_{e_1}^0, \mathbf{h}_{e_2}^0, \cdots, \mathbf{h}_{e_n}^0).
\end{equation}
The training objective is to predict the central entity $e$ with the input neighborhood context $\mathbf{c}_e=\text{MeanPooling}(\mathbf{h}_{e_1}^L,\mathbf{h}_{e_2}^L,\cdots,\mathbf{h}_{e_n}^L)$.
This can be regarded as a classification problem.
The prediction probability for each $e_i$ is given by
\begin{equation}
    \label{eq:ea_pro}
    p_{e_i} = \frac{\sigma(\mathbf{c}_e,\mathbf{h}_{e_i}^0)}{\sum_{e_j \in \mathcal{E}}\sigma(\mathbf{c}_e,\mathbf{h}_{e_j}^0)},
\end{equation}
where $\sigma()$ denotes inner product.
Let $\mathbf{q}_{e}$ be the one-hot label vector for predicting entity $e$. 
For example, $\mathbf{q}_{e}=[1,0,0,0,0]$ means that the entity at the first position is the target.
We adopt the label smoothing strategy to get soft labels, e.g., $\hat{\mathbf{q}}_{e}=[0.8,0.05,0.05,0.05, 0.05]$,
and use cross-entropy, i.e., $\text{CE}(x,y)=-\sum x \log y$, to calculate the loss between the soft label vector and the prediction distribution over all candidates $\mathbf{p}_{e}$. 
Formally, the overall loss is defined as
\begin{equation}
    \label{eq:neigh_loss}
    \mathcal{L}_\text{neighbor} = \sum_{e\in \mathcal{E}\cup\mathcal{E}'} \text{CE}(\hat{\mathbf{q}}_{e}, \mathbf{p}_{e}).
\end{equation}

\subsubsection{Cross-KG Path Encoding}
To capture the alignment information, we use seed entity alignment to bridge two KGs and perform random walks to generate cross-KG paths.
For example, a typical path of length $m$ can be denoted as $(e_1, e_2',...,e_m)$.
Then we randomly mask an entity in the path and get $(e_1, [MASK],...,e_m)$.
Similar to neighborhood encoding, we feed the masked path into Transformer to get its representations:
\begin{equation}
    \mathbf{h}_{e_1}^L, \mathbf{h}_{[MASK]}^L, \cdots, \mathbf{h}_{e_m}^L = \text{Encoder}(\mathbf{h}_{e_1}^0, \mathbf{h}_{[MASK]}^0, \cdots, \mathbf{h}_{e_m}^0).
\end{equation}
The output representation of $[MASK]$ is used to predict the masked entity $e_2'$. 
Here we use $\mathbf{h}_{[MASK]}^L$ to replace $\mathbf{c}_e$ in Eq. (\ref{eq:ea_pro}) to calculate the prediction probability.
Since the $[MASK]$ representation captures the information from all entities in the path, our model can capture the deep alignment interactions across two KGs.
If $e_2'$ has a counterpart $e_2$ in the seed entity alignment, 
our model also uses the $[MASK]$ representation to predict $e_2$, 
which means there are two labels for this prediction task.
The loss is given by:
\begin{equation}
    \label{eq:path_loss}
    \mathcal{L}_\text{path} = \sum_{e\in \mathcal{E}\cup\mathcal{E}'} \text{CE}(\hat{\mathbf{q}}_{e}, \mathbf{p}_{e}) + \sum_{(e, e')\in \mathcal{A}^{s}} \text{CE}(\hat{\mathbf{q}}_{e'}, \mathbf{p}_{e}).
\end{equation}

\subsubsection{Relation Regularization}
Relation triples are the most fundamental structural contexts of KGs, 
and have already demonstrated the effectiveness for entity alignment.
To encode relation triples into Transformer, we propose a relation regularization module at the input layer without increasing the model complexity and parameters.
The objective of regularization comes from the widely-used translational embedding model TransE \cite{bordes2013translating} that interprets a relation embedding as the translation vector from its subject entity to object entity.
Given a relation triple $(s, r, o)$, we expect the embeddings to hold that $\mathbf{h}_s^0 + \mathbf{r} = \mathbf{h}_o^0$, 
where $\mathbf{r}$ denotes the embedding of relation $r$. 
To realize such translation, we add a additional regularization loss by minimizing the translation error over all relation triples:
\begin{equation}
    \label{eq:relation_loss}
    \mathcal{L}_\text{relation} = \sum_{(s, r, o)\in \mathcal{T}\cup\mathcal{T}'}\| \mathbf{h}_s^0 + \mathbf{r} - \mathbf{h}_o^0\|_2.
\end{equation}

\subsection{Holistic Relation and Entity Alignment}
Given the learned embeddings, we propose a holistic reasoning method for the joint alignment of entities and relations. We use the input embeddings to conduct this process, i.e., $\mathbf{e} = \mathbf{h}_{e}^0$.
Different from existing methods that only consider entity embedding similarities to identify aligned entities, we also compare relations and neighbors to compute entity similarities.
The basic idea is that similar entities tend to have higher embedding similarity, and their relations and neighboring entities should also be similar. 

\subsubsection{Relation and Entity Functionality}
Intuitively, when comparing two entities, different relational neighbors have different importance. 
For example, relation \textit{birthplace} is more important than relation \textit{liveIn}, 
because a person can only be born in one place but can live in multiple places. 
Based on this, if we know two entities denote the same person, we can deduce that their neighbors along the relation \textit{birthplace} also indicate the same place. 
We use relation functionality \cite{suchanek2011paris} to quantify the importance of relations.
It measures the number of first arguments per relation-entity edge.
For example, given $r\in\mathcal{R}$, its functionality is calculated as:
\begin{equation}
    f(r) = \frac{|\{s|(s,r,o) \in \mathcal{T}\}|}{|(s, o)|(s,r,o) \in \mathcal{T}\}|},
\end{equation}
where $|\cdot|$ denotes the size of the set. Correspondingly, the inverse relation functionality is defined as:
\begin{equation}
    f^{-1}(r) = \frac{|\{o|(s,r,o) \in \mathcal{T}\}|}{|(s, o)|(s,r,o) \in \mathcal{T}\}|}.
\end{equation}

Analogously, we propose entity functionality for comparing two relations. It defines the importance of an entity in uniquely determining a relation. 
Rich connectivity of an entity would alleviate its importance. 
Therefore, we define entity functionality based on the number of relations an entity connects to, as a subject (functionality) or object (inverse functionality):
\begin{align}
    \begin{split}
    f(e) = \frac{1}{|\{(e,r,o)|(e,r,o) \in \mathcal{T}\}|}, \\ 
    f^{-1}(e) = \frac{1}{|\{(s,r,e)|(s,r,e) \in \mathcal{T}\}}.
\end{split}
\end{align}
All of these functionality features can be computed in advance, as they are only related to the statistics of KG connectivity.

\subsubsection{Relation Alignment Inference}
To assist entity alignment, we first find relation alignment in an unsupervised fashion. 
As each relation connects many entities, we utilize each relation's subject and object entity embeddings to approximate the relation embedding. 
Relations have the property of direction. Considering the heterogeneity of KGs that the relations from different KGs may be equivalent in a different direction, we generate embeddings for both relations and inverse relations by swapping the subject and object entity sets. 
The embeddings of connected entities are weighted by entity functionality. 
The approximated embeddings of each relation $\textbf{r}$ and its inverse $\textbf{r}_{inv}$ are calculated as follows:
\begin{equation}
    \mathbf{r} = [\mathbf{S}, \mathbf{O}],\,\,\,\mathbf{r}_{inv} = [\mathbf{O}, \mathbf{S}],
\end{equation}
where $\textbf{S}$ and $\textbf{O}$ represent the embedding of the subject entity set $\mathcal{E}_{\text{r}}^{\text{subj}}$ and object entity set $\mathcal{E}_{\text{r}}^{\text{obj}}$ of a relation $r$.
$[\cdot]$ denotes concatenation. 
The embeddings of the two sets can be generated by:
\begin{align}
    \begin{split}
        \textbf{S} = \sum_{s \in\mathcal{E}_{\text{r}}^{\text{subj}}} f(s) \times \mathbf{s},\,\,\,
        \textbf{O} = \sum_{o \in \mathcal{E}_{\text{r}}^{\text{obj}}}f^{-1}(o) \times \mathbf{o}
    \end{split}
\end{align}
Finally, we define the alignment score of a pair of relations as the cosine similarity of the approximated relation embeddings:
\begin{equation}
    \text{align}(r, r') = \cos(\mathbf{r}, \mathbf{r}').
\end{equation}
Here, $r$ can be either a relation or a reverse relation.

\subsubsection{Entity Alignment Inference}
After approximating the relation alignment score, 
we are able to match the relational neighbors (i.e., triples) of two entities from different KGs. 
Figure \ref{fig:relation_align.pdf} shows the process of measuring the equivalence between two entity neighborhood sets, which consists of the following steps:

\begin{figure}
    \centering
    \includegraphics[width=0.6\linewidth]{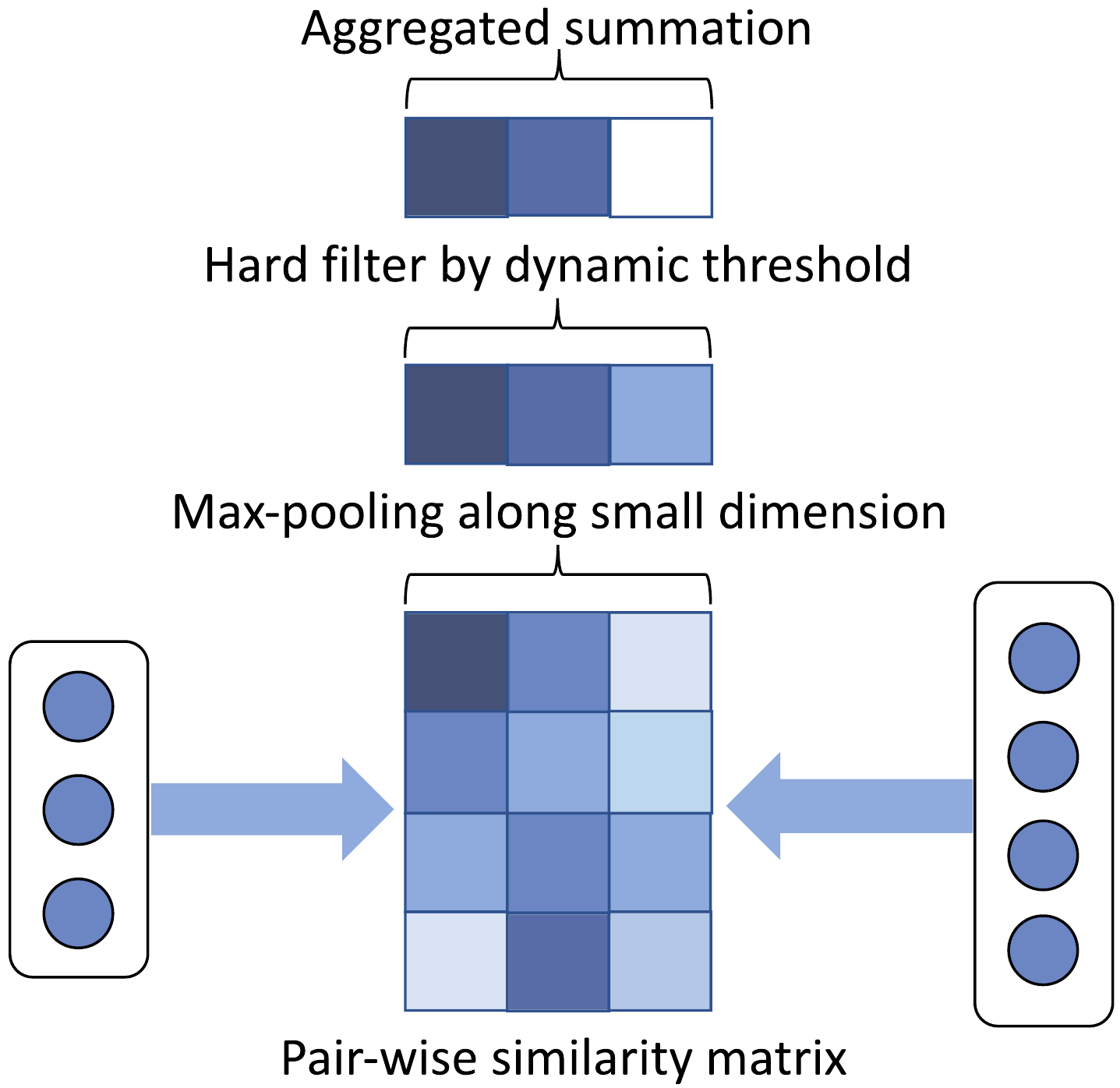}
    \caption{\label{fig:relation_align.pdf}Illustration of entity alignment inference.}
\end{figure}

\noindent\textbf{Step 1. Pair-wise matching matrix generation.}
Let $N_e$ and $N_{e'}$ denote the neighbourhood sets of $e$ and $e'$, respectively.
We first compute the pair-wise triple-level matching score for $e_i \in N_e$ and $e_j \in N_{e'}$.
We take into consideration both relation direction and functionality.
If the neighbour is outgoing, we utilize the inverse relation functionality $f^{-1}(r)$ for weighting; otherwise, we adopt its functionality $f(r)$. 
For brevity, we use the matching of two outgoing neighbors to illustrate the calculation of similarity matrix. 
Assuming the adjacent relations to the center entity of $e_i$ and $e_j$ are $r_i$ and $r_j$ respectively, we define $\phi$ as the similarity function of $e_i \in N_e$ and $e_j \in N_{e'}$ as follows:
\begin{equation}
    \phi(e_i,e_j) = f^{-1}(r_i) \times f^{-1}(r_j) \times \text{align}(r_i, r_j) \times \cos(\mathbf{e}_i,\mathbf{e}_j).
\end{equation}
\noindent\textbf{Step 2. Neighborhood intersection calculation.}
We then acquire the intersection of two neighborhood sets $N_e \cap N_{e'}$ based on the matching matrix. 
As the size of the smaller set indicates the maximum number of possibly matched entities that two sets may have, we conduct the max-pooling operation along the smaller dimension to identify possibly matched entities. 
Formally, assuming $|N_e| < |N_{e'}|$, the possibly matched entities are defined as:
\begin{equation}
    \Psi = \{(e_i, e_j)| i = 1,\cdots,|N_e|, j=\argmax_{j \in [1,|N_{e'}|]}\phi(e_i, e_j)\}.
\end{equation}
Then, we use a dynamic threshold $\eta$ to filter out impossible matches.
\begin{equation}
    \Theta = \{(e_i, e_j)| (e_i, e_j) \in \Psi,\ \phi(e_i, e_j) > \eta\}.
\end{equation} 
Finally, we sum up all the remaining similarities as the equivalence score of the two neighborhood sets.
This process can be regarded as computing the intersection of two sets.
In other words, the intersection score of $N_e$ and $N_{e'}$ is the aggregation of the triple-level scores from the intersection entity set:
\begin{equation}
    N_e \cap N_{e'} = \sum_{(e_i, e_j) \in \Theta}\phi(e_i, e_j).
\end{equation}

\noindent\textbf{Step 3. Neighborhood union calculation.} 
Next, we compute the union of neighborhoods to normalize the above equivalence score.
We aggregate the weighted edges after the max-pooling step.
Specifically, we use the relation functionality of connected edges as the weight and calculate the pair-wise functionality multiplication to approximate the union of two neighborhood sets. 
Taking outgoing edges as an example, for $e_i\in N_e$ and $e_j \in N_{e'}$, we have:
\begin{equation} 
    \phi^r(e_i, e_j) = f^{-1}(r_i)\times f^{-1}(r_j).
\end{equation}
Then, we aggregate the pair-wise functionality of possibly matched neighbors as the union score of two neighborhood sets $N_e \cup N_{e'}$:
\begin{equation}
    N_e \cup N_{e'} = \sum_{(e_i, e_j) \in \Psi}\phi^r(e_i, e_j).
\end{equation}

\noindent\textbf{Step 4. Holistic alignment probability. }
Finally, the entity alignment probability is calculated in a Jaccard-like fashion based on their neighborhoods, as defined below:
\begin{equation}
    \text{prob}(e, e') = \frac{N_e \cap N_{e'}}{N_e \cup N_{e'}}.
\end{equation}

\subsection{Soft Label Editing by Alignment Evidence}
After acquiring the holistic alignment probability of important entity pairs,
i.e., the top-k similar target counterparts of each source entity, 
we inform the multi-context Transformer of such an alignment clue to edit the soft labels.
We design two types of alignment evidence.
Specifically, we collect entity pairs with high alignment probability, i.e., $\text{prob}(e, e') \geq \gamma_{\text{pos}}$, as positive evidence.
On the contrary, if $\text{prob}(e, e') \leq \gamma_{\text{neg}}$, the two entities form a negative alignment pair.
Then, we inject both positive and negative alignment evidences into the embedding module by editing the soft labels for prediction tasks.
The idea is that if the target entity has several counterparts in positive pairs, all counterparts should also be labeled as $\min(s, p)$, where $s$ denotes the soft label of the target entity and $p$ is the calculated alignment probability.
We still use the soft label vector $\hat{\mathbf{q}}_{e}=[0.8,0.05,0.05,0.05,0.05]$ as an example, where the entity at the first position is the target. 
Suppose it has two predicted counterparts at the second and third positions with the probabilities of $0.5$ and $0.9$, respectively. In this case, the soft label vector is rewritten as $\hat{\mathbf{q}}_{e}=[0.8,0.5,0.8,0.05,0.05]$, and it is further normalized to serve as the labels for training the Transformer.
For negative counterparts, the corresponding label values are set to $0$ which will not be included in the prediction.

\section{Experiment}
\subsection{Experiment Setting}
\subsubsection{Datasets}
We use the benchmark dataset (V1) in OpenEA \cite{sun13benchmarking} for evaluation as it follows the data distribution of real KGs. 
It contains two cross-lingual settings extracted from multi-lingual DBpedia (English-to-French and English-to-German), and two monolingual settings among popular KGs (DBpedia-to-Wikidata and DBpedia-to-YAGO). 
Each setting has two sizes with 15K and 100K pairs of reference entity alignment, respectively, without reference relation alignment. 
We follow the data splits in OpenEA where $20\%$ alignment is for training, $10\%$ for validation and $70\%$ for testing. 

\subsubsection{Implementation Details}
Our \modelname model is implemented based on the OpenEA library. 
We initialize the trainable parameters with Xavier initialization \cite{Xavier} and optimize the loss with Adam \cite{Adam}.
As for hyper-parameters, the number of sampled neighbors is $3$, and the length of paths is $5$.
The batch size is $4096$, and the learning rate is $0.0001$.
The hidden size of Transformer is $256$. 
It has $8$ layers with $8$ attention heads in each layer. 
We conduct the soft label editing process based on holistic inference from the second training epoch.
The dynamic threshold of hard filter is $\eta = \max(0.6, 0.5 * (\max_{(e_i, e_j)\in \Theta}\phi(e_i, e_j) + \text{mean}_{(e_i, e_j)\in \Theta}\phi(e_i, e_j)))$. 
The soft label value is $s=0.6$. 
The thresholds for selecting positive and negative pairs are set as $\gamma_{\text{pos}} = 0.62$ and $\gamma_{\text{neg}} = 0.01$, respectively.
By convention, we use Hits@1, Hits@5 and mean reciprocal rank (MRR) as the evaluation metrics, and higher scores indicate better performance. 
We release our source code at GitHub\footnote{\url{https://github.com/JadeXIN/IMEA}}.

\begin{table*}[!t]
	\centering
	\caption{Entity alignment results on 15K datasets.}
	\resizebox{0.9\textwidth}{!}{
		\begin{tabular}{lcclcclcclccl}
			\toprule
			\multirow{2}{*}{Methods} & \multicolumn{3}{c}{EN-FR-15K} & \multicolumn{3}{c}{EN-DE-15K} & \multicolumn{3}{c}{D-W-15K} & \multicolumn{3}{c}{D-Y-15K} \\
			\cmidrule(lr){2-4} \cmidrule(lr){5-7} \cmidrule(lr){8-10} \cmidrule(lr){11-13} 
			& Hits@1 & Hits@5 & MRR & Hits@1 & Hits@5 & MRR & Hits@1 & Hits@5 & MRR & Hits@1 & Hits@5 & MRR \\ \midrule
			MTransE & 0.247 & 0.467 & 0.351 & 0.307 & 0.518 & 0.407 & 0.259 & 0.461 & 0.354 & 0.463 & 0.675 & 0.559 \\
			IPTransE & 0.169 & 0.320 & 0.243 & 0.350 & 0.515 & 0.430 & 0.232 & 0.380 & 0.303 & 0.313 & 0.456 & 0.378 \\
			AlignE & 0.357 & 0.611 & 0.473 & 0.552 & 0.741 & 0.638 & 0.406 & 0.627 & 0.506 & 0.551 & 0.743 & 0.636 \\
			GCN-Align & 0.338 & 0.589 & 0.451 & 0.481 & 0.679 & 0.571 & 0.364 & 0.580 & 0.461 & 0.465 & 0.626 & 0.536 \\
			SEA & 0.280 & 0.530 & 0.397 & 0.530 & 0.718 & 0.617 & 0.360 & 0.572 & 0.458 & 0.500 & 0.706 & 0.591 \\
			RSN4EA & 0.393 & 0.595 & 0.487 & 0.587 & 0.752 & 0.662 & 0.441 & 0.615 & 0.521 & 0.514 & 0.655 & 0.580 \\
			AliNet* & 0.364 & 0.597 & 0.467 & 0.604 & 0.759 & 0.673 & 0.440 & 0.628 & 0.522 & 0.559 & 0.690 & 0.617 \\
			HyperKA* & 0.353 & 0.630 & 0.477 & 0.560 & 0.780 & 0.656 & 0.440 & 0.686 & 0.548 & \underline{0.568} & \underline{0.777} & \underline{0.659} \\
			KE-GCN* & \underline{0.408} & \underline{0.670} & \underline{0.524} & \textbf{0.658} & \textbf{0.822} & \textbf{0.730} & \underline{0.519} & \underline{0.727} & \underline{0.608} & 0.560 & 0.750 & 0.644\\
			\midrule
			\modelname & \textbf{0.458} & \textbf{0.720} & \textbf{0.574} & \underline{0.639} & \underline{0.827} & \underline{0.724} & \textbf{0.527} & \textbf{0.753} & \textbf{0.626} & \textbf{0.639} & \textbf{0.804} & \textbf{0.712} \\
			\bottomrule
	\end{tabular}}
	\label{tab:ent_alignment_15k}
\end{table*}

\begin{table*}[!t]
	\centering
	\caption{Entity alignment results on 100K datasets.}
	\resizebox{0.9\textwidth}{!}{
	\begin{tabular}{lcclcclcclccl}
		\toprule
		\multirow{2}{*}{Methods} & \multicolumn{3}{c}{EN-FR-100K} & \multicolumn{3}{c}{EN-DE-100K} & \multicolumn{3}{c}{D-W-100K} & \multicolumn{3}{c}{D-Y-100K} \\
		\cmidrule(lr){2-4} \cmidrule(lr){5-7} \cmidrule(lr){8-10} \cmidrule(lr){11-13} 
		& Hits@1 & Hits@5 & MRR & Hits@1 & Hits@5 & MRR & Hits@1 & Hits@5 & MRR & Hits@1 & Hits@5 & MRR \\
			\midrule
			MTransE & 0.138 & 0.261 & 0.202 & 0.140 & 0.264 & 0.204 & 0.210 & 0.358 & 0.282 & 0.244 & 0.414 & 0.328 \\
			IPTransE & 0.158 & 0.277 & 0.219 & 0.226 & 0.357 & 0.292 & 0.221 & 0.352 & 0.285 & 0.396 & 0.558 & 0.474 \\
			AlignE & 0.294 & 0.483 & 0.388 & 0.423 & 0.593 & 0.505 & 0.385 & 0.587 & 0.478 & 0.617 & 0.776 & 0.691 \\
			GCN-Align & 0.230 & 0.412 & 0.319 & 0.317 & 0.485 & 0.399 & 0.324 & 0.507 & 0.409 & 0.528 & 0.695 & 0.605 \\
			SEA & 0.225 & 0.399 & 0.314 & 0.341 & 0.502 & 0.421 & 0.291 & 0.470 & 0.378 & 0.490 & 0.677 & 0.578 \\
			RSN4EA & 0.293 & 0.452 & 0.371 & 0.430 & 0.570 & 0.497 & 0.384 & 0.533 & 0.454 & 0.620 & 0.769 & 0.688 \\
			AliNet* & 0.266 & 0.444 & 0.348 & 0.405 & 0.546 & 0.471 & 0.369 & 0.535 & 0.444 & \underline{0.626} & \underline{0.772} & \underline{0.692} \\
			HyperKA* & 0.231 & 0.426 & 0.324 & 0.239 & 0.432 & 0.332 & 0.312 & 0.535 & 0.417 & 0.473 & 0.696 & 0.574\\
			KE-GCN* &  \underline{0.305} & \underline{0.513} & \underline{0.405} & \underline{0.459} & \underline{0.634} & \underline{0.541} & \textbf{0.426} & \textbf{0.620} & \textbf{0.515} & 0.625 & 0.791 & 0.700 \\
			\midrule
			\modelname & \textbf{0.329} & \textbf{0.526} & \textbf{0.424} & \textbf{0.467} & \textbf{0.641} & \textbf{0.549} & \underline{0.419} & \underline{0.614} & \underline{0.508} & \textbf{0.664} & \textbf{0.817} & \textbf{0.733}\\
			\bottomrule
	\end{tabular}}
	\label{tab:ent_alignment_100k}
\end{table*}

\subsubsection{Baselines}
To evaluate the effectiveness of our proposed \modelname model, we compare it with existing state-of-the-art supervised structure-based entity alignment methods. 
They can be roughly divided into the following categories.

\begin{itemize}
\item Triple-based models that capture the local semantics information of relation triples based on TransE, including MTransE \cite{chen2017multilingual}, JAPE \cite{sun2017cross}, AlignE \cite{sun2018bootstrapping} and SEA \cite{pei2019semi}.
\item Neighborhood-based models that use GNNs to exploit subgraph structures for entity alignment, including GCN-Align \cite{wang2018cross}, AliNet \cite{sun2020knowledge}, HyperKA \cite{sun2020knowledge}, KE-GCN \cite{yu2020generalized}.
\item Path-based models that explore the long-term dependency across relation paths, including IPTransE \cite{zhu2017iterative} and RSN4EA\cite{guo2019learning}.
\end{itemize}

Our model and the above baselines all focus on the structural information of KGs. For a fair comparison, we ignore other models that incorporate side information (e.g., attributes, entity names and descriptions) like RDGCN \cite{wu2019relation}, KDCoE \cite{chen2018co} and AttrGNN \cite{liu2020exploring}.

\subsection{Results and Analysis}
\subsubsection{Main Results}
Tables \ref{tab:ent_alignment_15k} and \ref{tab:ent_alignment_100k} present the entity alignment results on the OpenEA 15K and 100K datasets, respectively.
Results labelled by * are reproduced using the released source code with careful tuning of hyper-parameters, and the other results are taken from OpenEA \cite{sun13benchmarking}. 
We can see that \modelname achieves the \textbf{best} or \underline{second-best} performance compared with these SOTA baselines. 
It outperforms the best performing baselines on Hits@1 by $1\%-7\%$, except for the EN-DE-15K and D-W-100K datasets. 
For example, on 15K datasets, \modelname achieves a gain of $7.1\%$ by Hits@1 compared with HyperKA and $7.9\%$ against the latest method KE-GCN on D-Y-15K.
Overall, KE-GCN achieves the second-best performance. 
It exceeds our performance by 1.9\% on EN-DE-15K and 0.7\% on D-W-100K due to its strength in combining advanced knowledge embedding models with GCNs. 
However, the limited structure context of its embedding module prevents it from acquiring further improvement. 
On the 100K datasets, many baselines fail to achieve promising results due to the more complex KG structures and larger alignment space.
\modelname also acquires the best results on most large-scale datasets, except the second-best on D-W-100K, which demonstrates the practicability of our model. 
This is because \modelname can integrate multiple structural contexts, and the informed knowledge can further improve the performance by holistic reasoning. 
In summary, the results show the superiority of \modelname.

\subsubsection{Different Path Length}
\begin{figure*}
    \centering
    \includegraphics[width=0.9\linewidth]{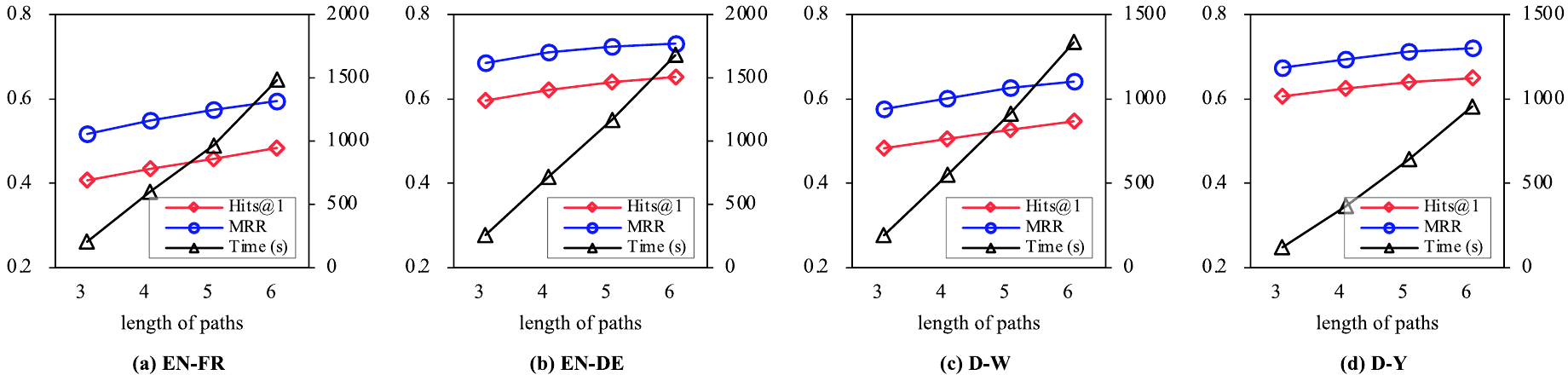}
    \caption{\label{fig:path_len}Performance and training time w.r.t path length on 15K datasets.}
\end{figure*}
As a path is one of the crucial structural contexts for both embedding and alignment learning, we evaluate the effect of path length on entity alignment. 
Figure \ref{fig:path_len} reports the \modelname performance w.r.t. different path lengths on 15K datasets. 
As expected, entity alignment accuracy (Hits@1 and MRR) improves as the path length increases.
This is because a longer path can boost the cross-KG interaction and mitigate the heterogeneity issue between different KGs. 
The performance increase is more evident on EN-FR-15K due to the higher structural heterogeneity in this dataset. 
However, it is also noticeable that the training time per epoch increases as the path length grows. 
Hence, we choose a path length of 5 in our setting as an effectiveness-efficiency trade-off.

\subsubsection{Complementarity of Different Context}

\begin{table}[!t]
\centering
\caption{Complementarity of context on 15K datasets.}
\resizebox{0.99\linewidth}{!}{
\begin{tabular}{lcccccccc} 
\toprule
\multirow{2}{*}{Methods} & \multicolumn{2}{c}{EN-FR} & \multicolumn{2}{c}{EN-DE} & \multicolumn{2}{c}{D-W} & \multicolumn{2}{c}{D-Y} \\
\cmidrule(lr){2-9}
& Hits@1 & MRR & Hits@1 & MRR & Hits@1 & MRR & Hits@1 & MRR \\
\midrule
\modelname & 0.458 & 0.574 & 0.639 & 0.724 & 0.527 & 0.626 & 0.639 & 0.712 \\ 
\,\, w/o relation & 0.378 & 0.482 & 0.609 & 0.684 & 0.460 & 0.552 & 0.599 & 0.667\\
\,\, w/o neighbor & 0.431 & 0.547 & 0.612 & 0.681 & 0.494 & 0.595 & 0.635 & 0.708\\
\bottomrule
\end{tabular}}
\label{tab:context-effect}
\end{table}

Table~\ref{tab:context-effect} gives the results of the ablation study on the triple and neighborhood contexts in 15K datasets (paths are necessary for alignment learning).
\modelname ``w/o relation" and ``w/o neighbor" represent the variants by removing relation triples or neighborhood information from our multi-context Transformer. 
The performance of both variants drops. 
When the relation triple context is removed, the Hits@1 decreases by $4\%$ to $8\%$, and MRR drops by $4\%$ to $9\%$. 
The decline is smaller when the neighborhood is absent, which is $0.5\%$ to $3\%$ drop on Hits@1 and $0.5\%$ to $4\%$ drop on MRR. 
Overall, the absence of both relation and neighbor contexts results in a performance drop on all datasets, and relation triples contribute more than neighborhood.

\subsubsection{Effect of Informed Process}
The ablation study on the effect of the holistic reasoning process is shown in Table \ref{tab:inf-effect}, 
where \modelname (w/o inf) represents the variant without using the referred holistic knowledge. It can be observed that the alignment clue captured by holistic inference can bring improvement on all datasets by $1\%$ to $2.5\%$ on Hits@1 and $1\%$ to $2\%$ on MRR. 
This quantitatively validates the effectiveness of our holistic reasoning process and its ability to inform embedding and alignment learning.

\begin{table}[!t]
\centering
\caption{Effect of holistic inference on 15K datasets.}
\resizebox{0.99\linewidth}{!}{
\begin{tabular}{lcccccccc} 
\toprule
\multirow{2}{*}{Methods} & \multicolumn{2}{c}{EN-FR} & \multicolumn{2}{c}{EN-DE} & \multicolumn{2}{c}{D-W} & \multicolumn{2}{c}{D-Y} \\
\cmidrule(lr){2-9}
& Hits@1 & MRR & Hits@1 & MRR & Hits@1 & MRR & Hits@1 & MRR \\
\midrule
\modelname & 0.458 & 0.574 & 0.639 & 0.724 & 0.527 & 0.626 & 0.639 & 0.712 \\ 
\,\, w/o inf & 0.431 & 0.546 & 0.614 & 0.703 & 0.502 & 0.598 & 0.629 & 0.704\\
\bottomrule
\end{tabular}}
\label{tab:inf-effect}
\end{table}

\subsection{Case Study on Holistic Reasoning}
In this section, we conduct case studies to further demonstrate the effectiveness of our holistic reasoning, i.e., how the positive and negative evidence obtained from holistic reasoning can be used to fix the alignment error caused by learned embeddings.

\begin{figure}[t]
    \centering
    \begin{subfigure}[b]{0.95\linewidth}
    \centering
    \includegraphics[width=1\linewidth]{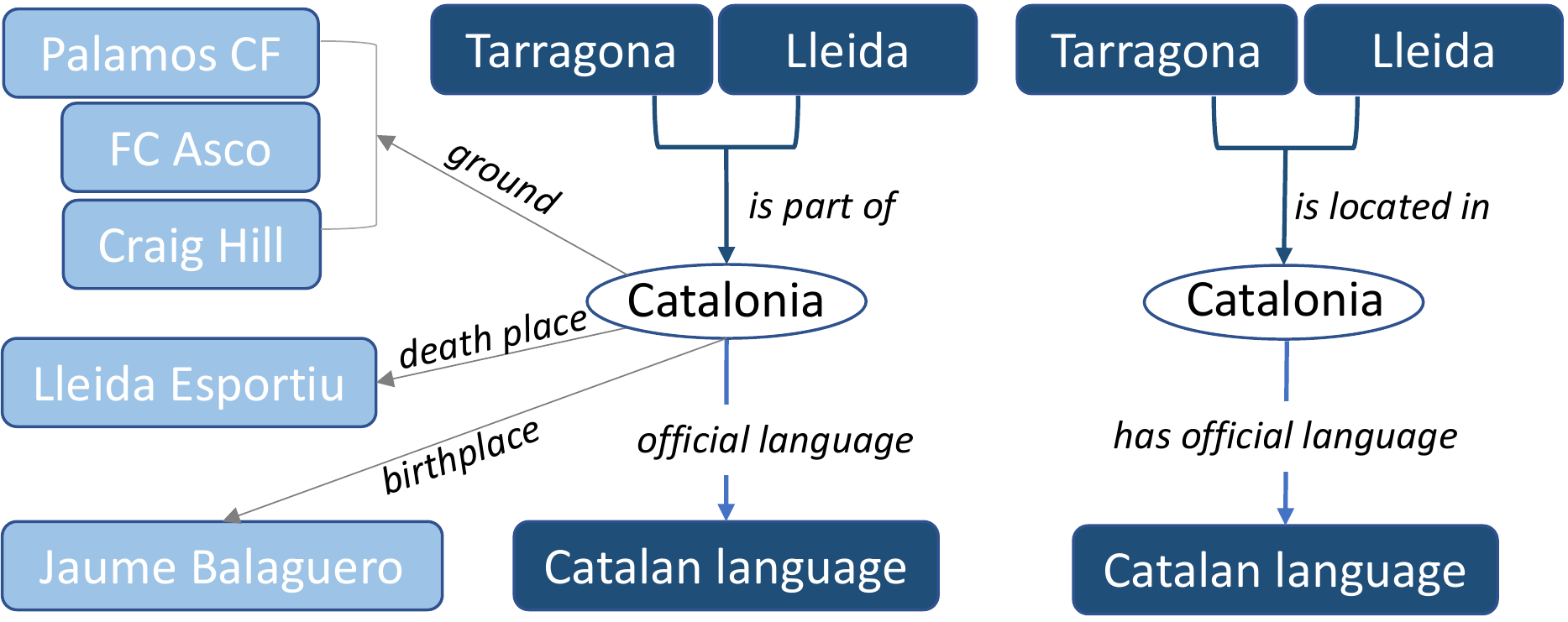}
    \caption{\label{fig:case2.pdf}}
    \end{subfigure}
    \begin{subfigure}[b]{0.95\linewidth}
    \includegraphics[width=1\linewidth]{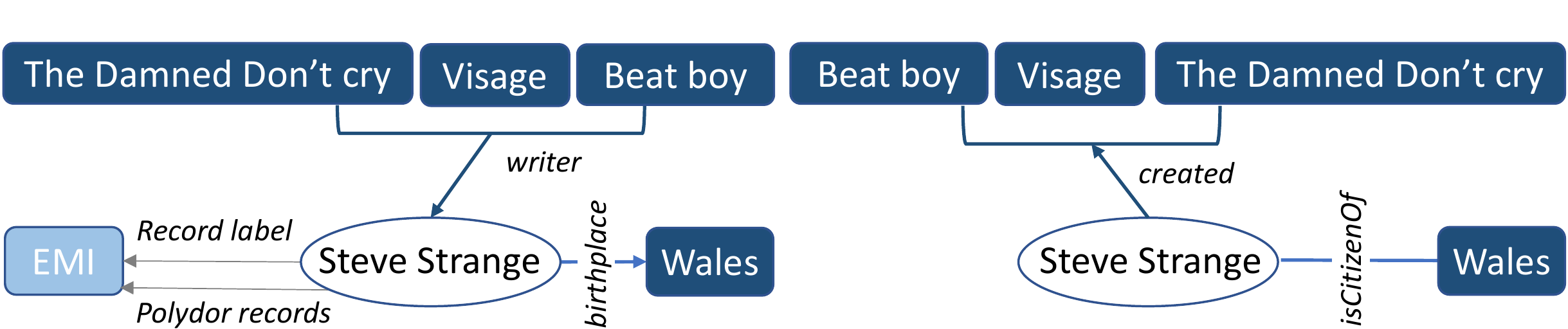}
    \caption{\label{fig:case3.pdf}}
    \end{subfigure}
    \caption{Examples of aligned entity pairs discovered by holistic inference from D-Y-15K. The dark arrow and dark rectangle indicate the matched relations and entities recognized by the inference, respectively.}
    \label{fig:correct example}
\end{figure}

\begin{figure}[t]
    \centering
    \begin{subfigure}[b]{0.5\textwidth}
    \centering
    \includegraphics[width=1\linewidth]{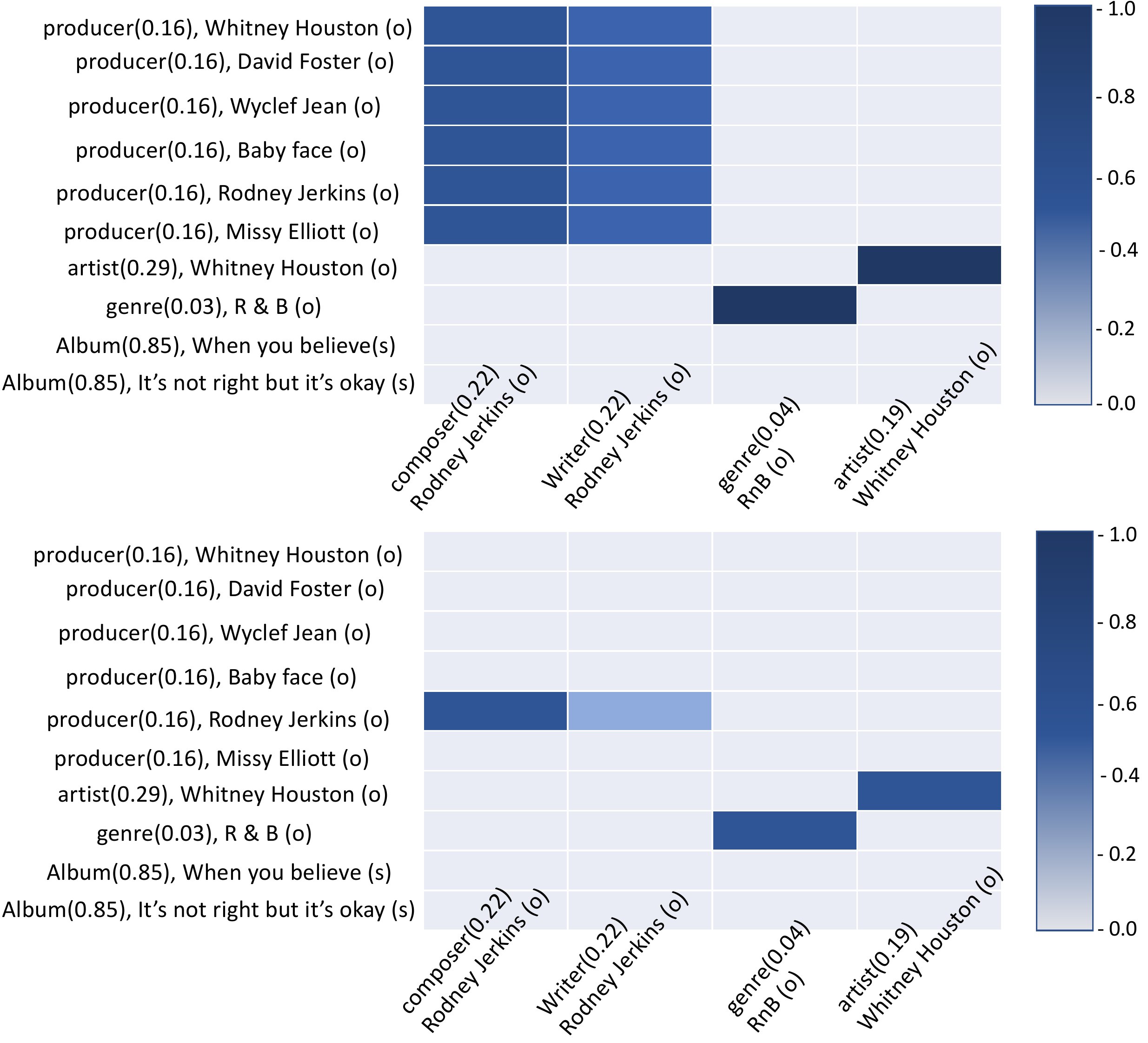}
    \caption{\label{fig:case1_0.pdf}Incorrect alignment pairs.}
    \end{subfigure}
    \begin{subfigure}[b]{0.5\textwidth}
    \centering
    \includegraphics[width=1.0\linewidth]{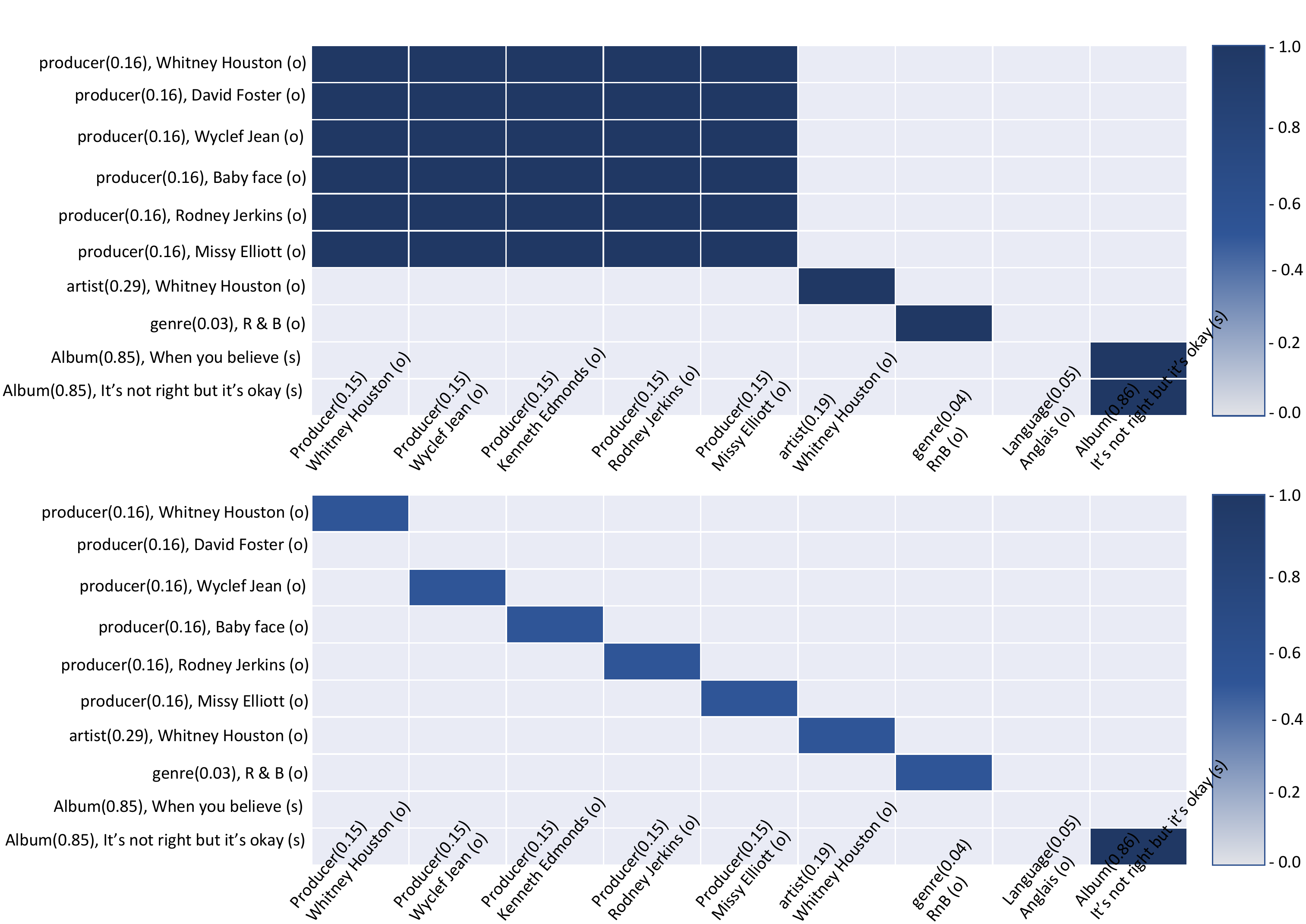}
    \caption{\label{fig:case1_1.pdf}Correct alignment pairs.}
\end{subfigure}
\caption{Example of a negative alignment from EN-FR-15K which can be corrected by the holistic reasoning. 
The vertical and horizontal axis represents the entity's neighbors from EN and FR respectively. Each neighbor consists of the relation (with its functionality) and the entity (either subject "s" or object "o"). The top figure reports the relation alignment while the bottom figure shows the neighbor alignment. Darker cells indicate higher alignment probability between the corresponding relations or neighbors.
}
\label{fig: incorrect example}
\end{figure}

\subsubsection{Positive Evidence Benefit Embedding Learning}
Figure \ref{fig:case2.pdf} illustrates a case of positive alignment evidence from D-Y-15K. 
The learned embedding similarity between these two identical entities is only $0.59$. 
Such an inaccurate embedding distance is caused by the sparse connections between the two entities. 
Long-tail entities suffer from the isolation issue. Hence, the two entities with sparse triples would have few chances to be pushed closer during training. 
Another reason for the low embedding similarity is the large degree difference between the two entities, since the embedding model tends to generate dissimilar representations for entities with different degrees.
Fortunately, our holistic inference can effectively capture the positive interaction between entities by matching their neighborhood subgraphs at a more fine-grained level (i.e., triple-level). 
Firstly, it can catch the matched relations, i.e., \textit{``is-part-of''} and \textit{``is-located-in''}, \textit{``official-language''} and \textit{``has-official-language''}. 
As the relational neighbors of the two entities are not always long-tail entities, the matched neighbors (denoted by dark blue) can be successfully captured, which leads to a positive evidence score of $0.92$.
Figure \ref{fig:case3.pdf} demonstrates another example from D-Y-15K, which shows the holistic process is especially suitable for handling the heterogeneous issue between KGs. 
From our observation, even though two entities have many relational neighbors in common, the generated embeddings tend to be more dissimilar if the relations are in different directions.
The embedding similarity in Figure \ref{fig:case3.pdf} is only $0.56$. 
As the holistic process considers relation direction, it successfully captures equivalent relations \textit{``writer''} and \textit{``created''} with different directions, and similar relations \textit{``birthplace''} and \textit{``is-citizen-of''}, 
which leads to a high inference score of $0.89$.
Overall, for those equivalent entities with low embedding similarity,
our holistic inference can successfully capture their actual alignment interactions by comparing their neighborhood subgraphs, thus fixing the error caused by unreliable embedding learning.

\subsubsection{Negative Evidence Benefit Embedding Learning}
Our holistic reasoning can also capture the negative alignment evidence between some wrongly closely located entity embeddings, and correctly accumulate positive clues between matched entities. 
Figure \ref{fig: incorrect example} shows an example from EN-FR-15K about two songs from ``Whitney Houston'', where the incorrectly aligned entity pairs in Figure \ref{fig:case1_0.pdf} have a slightly higher embedding similarity ($0.99$) than the corresponding ground-truth in Figure \ref{fig:case1_1.pdf} ($0.97$). Such an embedding mistake can be corrected by our holistic reasoning based on the neighborhood information.
The neighboring entities' embeddings could also be inaccurate, which results in a closer embedding distance between the entity and its incorrect match. 
However, by holistic inference, we observe that the incorrectly aligned entities in Figure \ref{fig:case1_0.pdf} only share some unimportant neighbors with low relation functionality such as \textit{``genere''}, \textit{``artist''}, \textit{``producer''} and \textit{``composer''}, leading to a limited alignment score of $0.32$. On the contrary, the ground-truth alignment in Figure \ref{fig:case1_1.pdf} demonstrates a large number of common neighbors, among which positive matching clues are observed on some crucial neighbours with large relation functionality like \textit{``album''} (the corresponding entities \textit{``It's not right but It's okay''} are matched). Hence, it accumulates more positive evidences and achieves a higher alignment probability of $0.76$. In this way, holistic inference successfully identifies the aligned entity pair in Figure \ref{fig:case1_0.pdf} as a negative prediction from the embedding module (with a smaller alignment probability), and injects such feedback into the Transformer to adjust embedding learning, thus improving the final entity alignment performance.

\section{Conclusion and Future Work}
In this paper, we propose an informed multi-context entity alignment model (\modelname). It consists of a multi-context Transformer to capture multiple structural contexts for alignment learning, and a holistic reasoning process to generate global alignment evidence, which is iteratively injected back into the Transformer via soft label editing to guide further embedding training. Extensive experiments on the OpenEA dataset verify the superiority of \modelname over existing state-of-the-art baselines. As future work, we will include extra side information of KGs to enhance embedding learning in \modelname, and generate human-understandable interpretations of entity alignment results based on the informed holistic reasoning, thus achieving explainability of embedding-based EA models.

\begin{acks}
This work was partially supported by the Australian Research Council under Grant No. DE210100160 and DP200103650.
Zequn Sun's work was supported by Program A for Outstanding PhD Candidates of Nanjing University.
\end{acks}

\bibliographystyle{ACM-Reference-Format}
\balance
\bibliography{sample-base}

\end{document}